\def\year{2020}\relax
\documentclass[letterpaper]{article} 
\usepackage{aaai20}  
\usepackage{times}  
\usepackage{helvet} 
\usepackage{courier}  
\usepackage[hyphens]{url}  
\usepackage{graphicx} 
\usepackage{graphicx}  
\usepackage[utf8]{inputenc}
\usepackage{amsmath}
\usepackage{booktabs}
\usepackage{algorithm}
\usepackage{algorithmic}
\usepackage{subfigure}

\usepackage{amssymb}
\usepackage{amsthm}
\usepackage{tikz}
\usetikzlibrary{positioning}
\title{A Generalized Reinforcement Learning Algorithm for Online 3D Bin-Packing}
\author{\Large{\textbf{Richa Verma\textsuperscript{\rm 1}, Aniruddha Singhal\textsuperscript{\rm 1}, Harshad Khadilkar\textsuperscript{\rm 1}, Ansuma Basumatary\textsuperscript{\rm 1}}}\\\Large{\textbf{Siddharth Nayak\textsuperscript{\rm 2}, Harsh Vardhan Singh\textsuperscript{\rm 1}, Swagat Kumar\textsuperscript{\rm 1} and Rajesh Sinha\textsuperscript{\rm 1}}}\\
\textsuperscript{\rm 1}Tata Consultancy Services Research and Innovation Lab, Thane, India,\\\textsuperscript{\rm 2}Indian Institute of Technology Madras, Chennai, India\\
\textsuperscript{\rm 2}\{\textrm{siddharthnayak98\}@gmail.com} 
}
 
\begin{document}

\maketitle

\begin{abstract}
We propose a Deep Reinforcement Learning (Deep RL) algorithm for solving the online 3D bin packing problem for an arbitrary number of bins and any bin size. The focus is on producing decisions that can be physically implemented by a robotic loading arm, a laboratory prototype used for testing the concept. The problem considered in this paper is novel in two ways. First, unlike the traditional 3D bin packing problem, we assume that the entire set of objects to be packed is not known a priori. Instead, a fixed number of upcoming objects is visible to the loading system, and they must be loaded in the order of arrival. Second, the goal is not to move objects from one point to another via a feasible path, but to find a location and orientation for each object that maximises the overall packing efficiency of the bin(s). Finally, the learnt model is designed to work with problem instances of arbitrary size without retraining. Simulation results show that the RL-based method outperforms state-of-the-art online bin packing heuristics in terms of empirical competitive ratio and volume efficiency.
\end{abstract}

\section{Introduction} \label{sec:intro}

The geometric three-dimensional bin-packing problem is a known NP-hard problem in computer science literature, and is a variant of the knapsack problem \cite{Kolhe2010}. The goal is to define a packing strategy for boxes of different shape and size such that the space inside the bin is maximally utilized (volume efficiency version of the objective). Alternatively, one can define the goal in terms of the least number of bins for packing a given set of objects (competitiveness ratio version of the objective). In this paper, we describe an instantiation of the 3D bin-packing problem with the example of parcel loading using a robotic arm. The parcels (boxes) arrive on a conveyor belt, and only a subset of the entire batch is visible at any one time. This version of the problem is thus \textit{online} or \textit{real-time} 3D bin-packing, which we abbreviate to RT-3D-BPP. Since the concept is developed on a laboratory setup including the robotic arm and a conveyor, we include realistic constraints on the loading plan, including bottom-up ordered placement of boxes in the container, smoothness of the surface underneath each box, and degrees of freedom of the arm. 

A number of methods have been reported in literature for solving the standard (offline) 3D-BPP problem, where the complete set of objects to be packed is known a priori. This includes integer linear programming \cite{den2003note}, space minimizing heuristics \cite{Crainic2007}, genetic algorithms \cite{Goncalves2013}, and machine learning based smart heuristic selection \cite{Ross:2002:HLC:2955491.2955659}. Compared to the offline version, RT-3D-BPP is more challenging as the packing strategy is decided for each arriving object with limited knowledge of the future sequence of objects. Hence, one can only hope to locally optimize the policy, which is itself complicated by the dynamically changing fitness landscape \cite{weise2009optimization} and the limited computation time. Most existing online methods focus on finding the theoretical bounds for approximate or heuristic methods for online bin packing \cite{epstein2009online}, \cite{han2011new}, \cite{christensen2017approximation}, \cite{epstein2007bounds}. Heuristic methods require a large amount of design effort and have restricted generalization capabilities. By contrast, Deep Reinforcement learning (Deep RL) has shown promise in solving several combinatorial problems \cite{DBLP_SutskeverVL14}, \cite{DBLP_BelloPLNB16}. Furthermore, it has also been used for long horizon planning problems such as in the case of Alpha Go \cite{silver2017mastering}. 

While Deep RL has recently been applied to the bin-packing problem as well \cite{jin2017deep}, \cite{hu2017solving}, prior literature focuses on the offline version of the problem. In this paper, we propose a novel Deep RL method for solving the RT-3D-BPP. The proposed method uses a DQN framework to learn suitable packing strategies while taking practical robotic constraints into account. We compare the performance of the RL methodology against state-of-the-art heuristics. The metric used for comparison is the empirically computed competitive ratio, a measure of the number of bins required by an algorithm to pack a set of objects, normalized by the optimal number of bins. While the RL algorithm is trained using a simulator, the practicability of the trained policy is tested through a laboratory experiment with physical robots, and on problem instances with different scale and underlying distribution of object dimensions.

The claimed contributions of this paper are, (1) a novel heuristic (called WallE) for solving RT-3D-BPP which is shown to outperform existing bin-packing heuristics, (2) a Deep RL methodology (called PackMan) for online computation of object location and orientation, combined with (3) a generalised approach that allows the algorithm to work with arbitrary bin sizes, making it more suitable for real world deployment. The rest of this paper is organized as follows. An overview of related work is provided in the next section, followed by a formal problem definition, proposed methods for solving RT-3D-BPP, and details of various simulation and real world experiments and their outcomes.


\section{Related Work} \label{sec:related}

We briefly review literature dealing with the real-time and offline 3D bin-packing problems (which also gives us the performance metric used here), followed by reinforcement learning solutions to combinatorial problems in general, and finally learning-based methodologies for bin-packing.

\textbf{Solutions to bin-packing problems:} 
The online version of the 3D bin-packing problem (RT-3D-BPP) is commonly found in the warehouse parcel loading context, and has been formulated as a knapsack problem \cite{Kolhe2010}. The survey \cite{Kolhe2010} is limited to the description of the problem and benchmarks that are used to compare different algorithms. A more detailed study of n-dimensional BPP \cite{Christensen2016} describes state-of-the-art algorithms and their `competitive ratio'. 

\textit{\textbf{Definition}: An online algorithm $A$ is called $c$-competitive if there exist constants $c$ and $\delta$ such that for all finite input sequences $I$, the number of bins required by $A$ is upper-bounded by $c\cdot Opt(I) + \delta$, where $Opt(I)$ is the optimal number of bins for this input sequence. }

Online algorithms with $c=1.58$ are known for rectangle and cube packing with and without rotation \cite{balogh2017new}. An alternative metric for bin-packing is the percentage of the filled space inside the container. While we report these numbers for our algorithms in this paper, we did not find an existing comparison of algorithms based on this metric. Also, to the best of our knowledge, there are no standard data sets on which performance of different algorithms can be tested empirically.
	
RT-3D-BPP becomes challenging when the problem is posed as placement through an articulated robotic arm. Robotic arms may have  limits on reachability inside the container and limited manoeuvrability. They cannot reshuffle the objects which are placed inside the container and cannot undertake very tight placements in context of sensor noise and minor caliberation errors. 3D-BPP for robotic arm is discussed in \cite{martello2007algorithm,den2003note}, but there is no discussion of the online variant. A more recent and advanced study for irregular objects has been done in \cite{wang2018stable}. Study of column formation technique for robot arm packaging is discussed in \cite{Mahvash2017}.  A system for packing homogenous sized boxes using robotic arm is discussed in \cite{justesen2018automated}.
	
\textbf{Reinforcement learning for combinatorial optimization: }
Recent literature has demonstrated the potential of reinforcement learning (RL) for solving combinatorial optimization problems, with applications to gameplay \cite{silver2017mastering}, operations research problems such as travelling salesperson (TSP) \cite{gambardella1995252,DBLP_BelloPLNB16}, vehicle routing (VRP) \cite{nazari2018reinforcement} and job-shop scheduling (JSSP) \cite{zhang_1643031_1643044,wei2005reinforcement}, and logistics problems such as container loading for ships \cite{verma2019reinforcement}. Apart from gameplay situatons, most combinatorial optimization problems can be formulated as one-shot scheduling problems. However, for comptability with RL algorithms, they are reformulated as sequential decision-making problems. We follow a similar conceptual approach, by treating each parcel loading step as a decision. However, there are three key differences between the current problem and ones considered earlier. First, unlike in gameplay situations, the number of potential actions for our problem (location and orientation of boxes) is very large. Second, unlike TSP and VRP, the change in state of the system after loading a box or parcel is complex, because it changes the surface shape of the container in 3D. Finally, the `goodness' of a chosen location can change substantially with very small changes in the action. For example, placing a box flush with the container wall is significantly better than leaving a small gap with the wall. As a result, RL algorithms with continuous action spaces such as DDPG \cite{ddpg_modified} are difficult to use. A more detailed discussion of algorithm selection is provided in subsequent sections.

	
	
\textbf{Supervised and reinforcement learning for bin-packing: }
A few studies have used heuristics augmented by supervised learning for bin-packing. In \cite{Ross:2002:HLC:2955491.2955659}, the authors have used a classifier system called XCS to select a heuristic from a given set, based on the percentage of boxes left to packed, to solve a 1D-BPP. Similarly,  \cite{DBLP:journals/corr/MaoBFJGMYGKT17} has used neural networks to select the heuristic based on feature information collected from the items and the bins. A few reinforcement learning based approaches have also been reported. In \cite{DBLP:journals/corr/abs-1807-01672}, authors apply their Reward Ranked (R2) algorithm to 2D and 3D BPP. This R2 algorithm computes ranked rewards by comparing the terminal reward of the agent against its previous performance, which is then used to update the neural network. In \cite{Runarsson_2177360_2177392}, the author proposes an RL algorithm which tries to learn heuristic policies using BPP as an example. Deep RL has been used in designing a bin with least surface area that could pack all the items \cite{DBLP_abs_1708_05930}, which uses policy-based RL (Reinforce)  with a 2-step Neural Network (Ptr-Net) consisting of RNNs and has shown to achieve $\approx 5\%$ improvement over heuristics. The authors improve the algorithm in \cite{DBLP_abs_1804_06896} by multi-tasking the sequence generation \cite{DBLP_SutskeverVL14}. 

Our survey did not reveal learning-based approaches that could solve the online 3D bin-packing problem, with high solution quality and limited computational time, and with real-world physical constraints on the loading system. In the next section, we describe the RT-3D-BPP context in detail, and proceed to develop an RL approach to solve this problem. The generalization capability of this algorithm is demonstrated in the Results section.
	
\section{Problem Description} \label{sec:problemdescription}

\begin{figure*}[!th]
\centering
\includegraphics[width=0.32\linewidth]{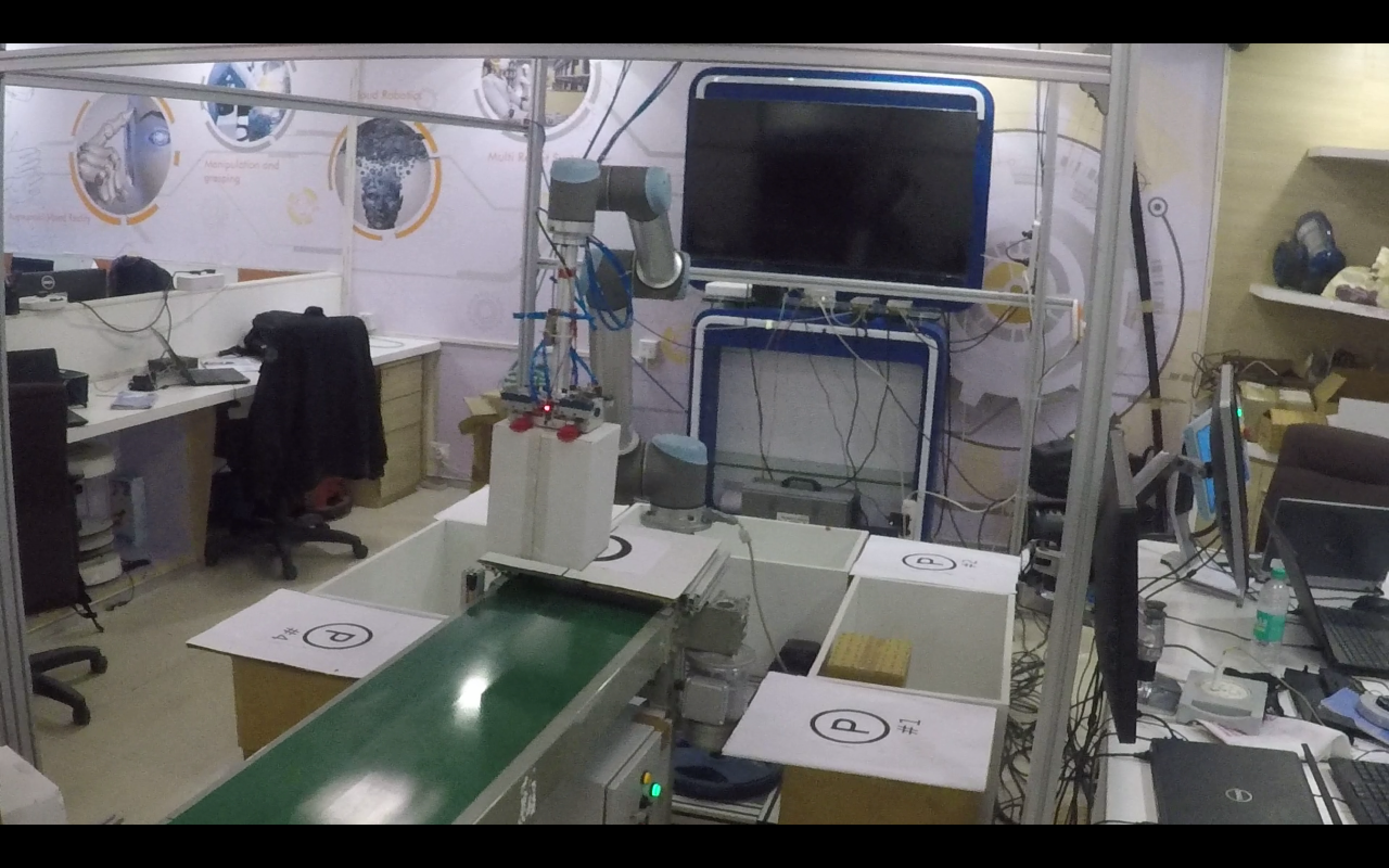}
\includegraphics[width=0.32\linewidth]{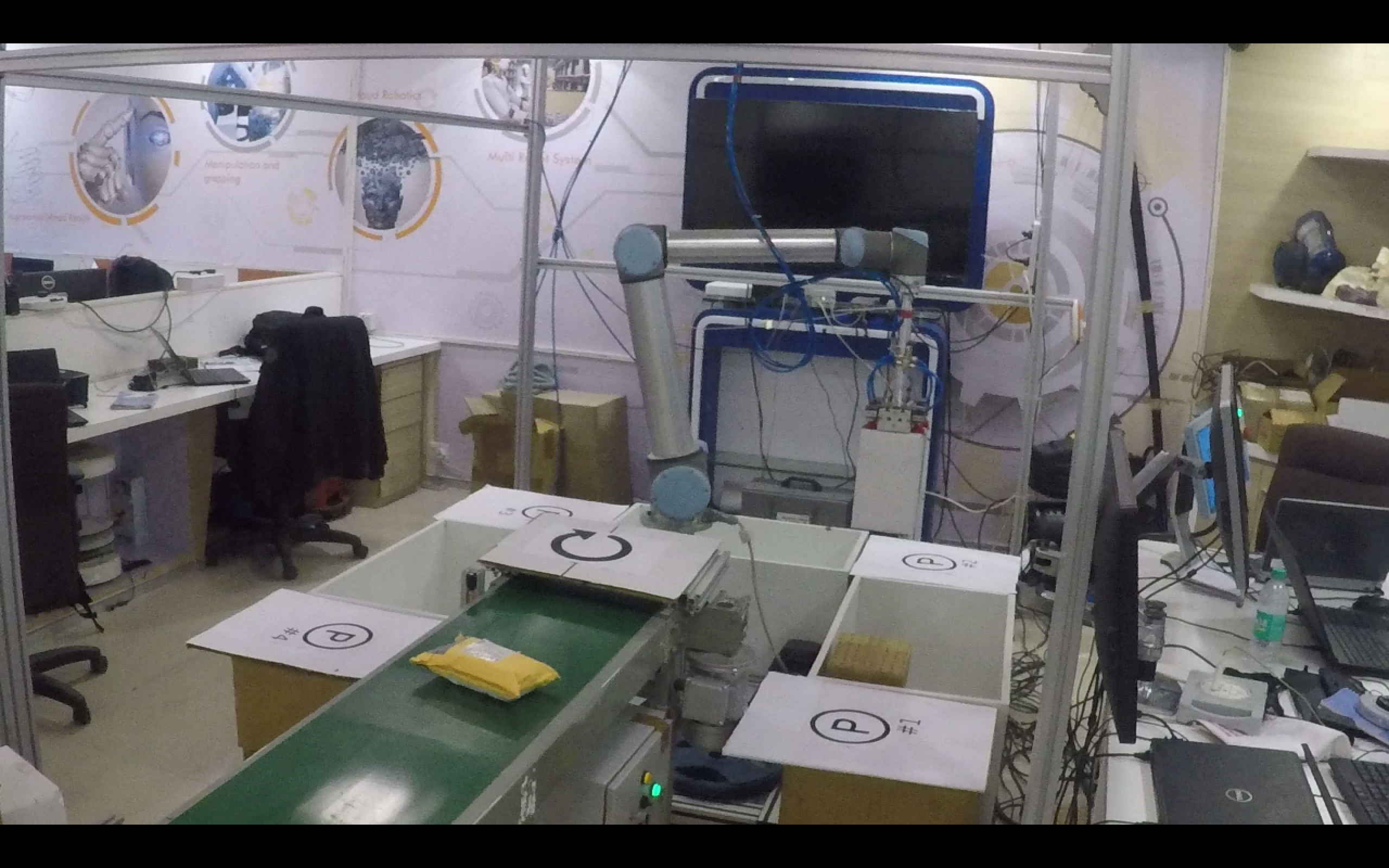}
\includegraphics[width=0.32\linewidth]{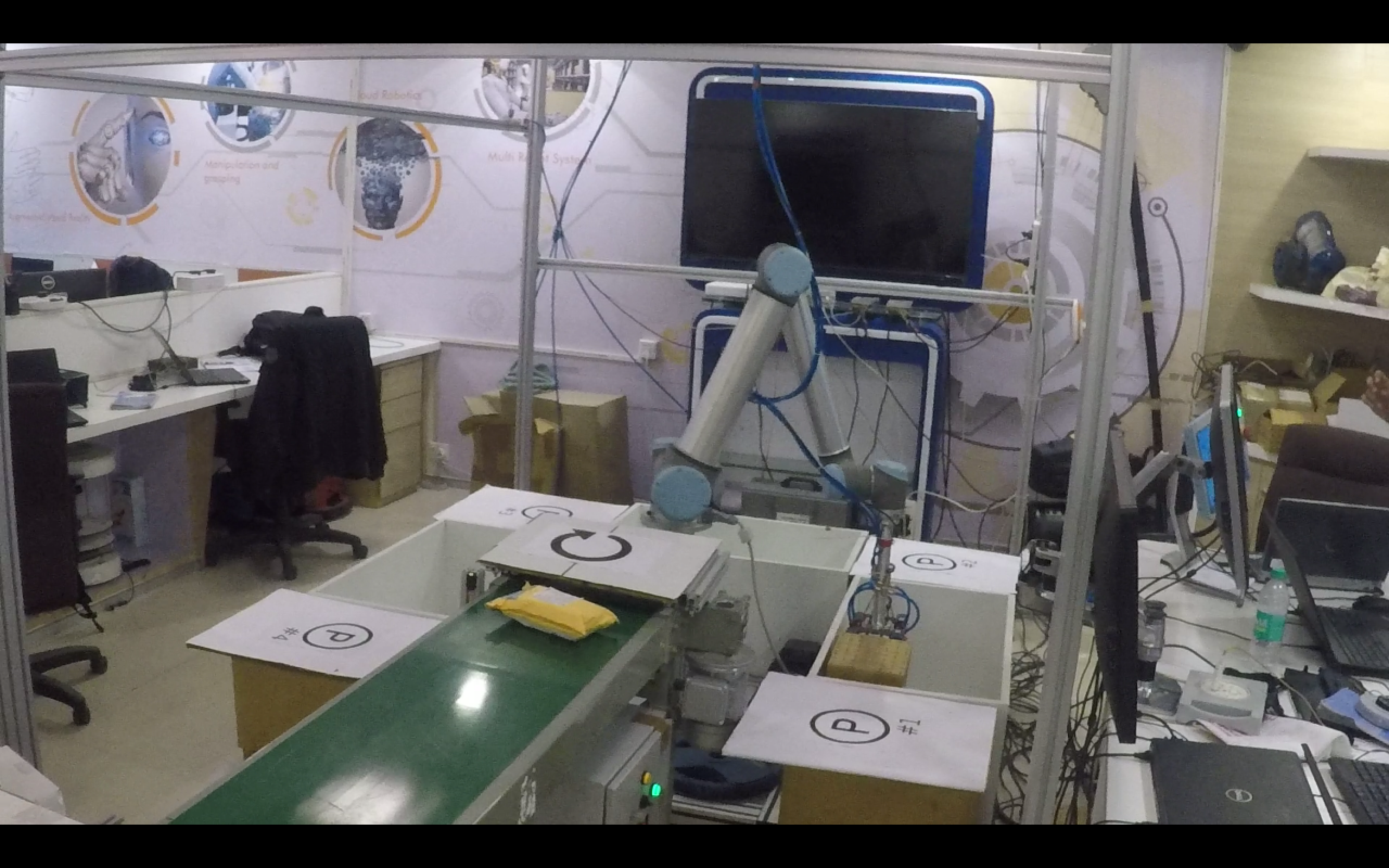}
\caption{Prototype created in the lab. The conveyor belt brings single-sorted parcels with regular spacing. Demonstration of one pick and place cycle is shown, from left to right (a) Picking the parcel from the conveyor belt (b) Searching for the placement location (c) Placing the parcel inside the container.}
\label{fig:lab_photo}
\end{figure*}

The overarching goal of this work is to develop a planning algorithm for a robotic parcel loading system in a sorting center. A mock-up of this system has been created in the laboratory, as shown in Fig. \ref{fig:lab_photo}. The setup is meant to test the robot-implementability of any developed packing policies, including constraints that are specific to the system being used (described in detail later). The equipment consists of a conveyor belt on which cuboidal parcels of arbitrary dimensions appear as a stream. We assume that the incoming parcels are ordered and equally spatially separated on the belt. Each parcel is picked by the robot arm and placed inside the container (Fig. \ref{fig:lab_photo}). The placement location is selected by the algorithm which is discussed in this paper. 

\begin{figure*}[!h]
\centering
\subfigure[Rotation: single axis]{\includegraphics[width=0.20\linewidth]{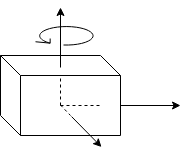}}
\subfigure[Placement: bottom-up]{\includegraphics[width=0.25\linewidth]{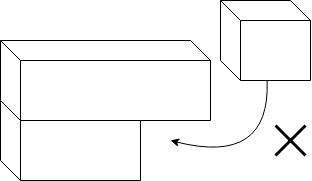}}
\subfigure[Flat base]{\includegraphics[width=0.12\linewidth]{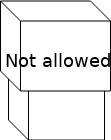}}
\subfigure[Round-off error]{\includegraphics[width=0.25\linewidth]{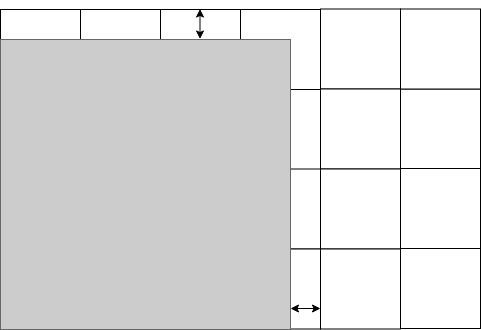}}
\caption{Restrictions due to robot packability constraints and discrete state space representation.}
\label{fig:assumptions}
\end{figure*}

\textbf{\textit{Robot-implementability:}} The loading of parcels is subject to some physical constraints. First, the gripper of the robot requires all boxes to be non-deformable cuboids. Second, the robot is capable of rotating the boxes only along the z-axis (vertical axis), in steps of 90$^o$. This is more restrictive than the generic RT-3D-BPP with six canonical orientations and partial rotations. Third, the robot placement has an accuracy of 1 centimetre, with any partial dimensions rounded up to the next highest centimetre. Fourth, the placement of sensors dictates that only the upcoming $n$ boxes (where $n$ is a parameter) are known to the RL agent, in terms of physical dimensions. Finally, parcels cannot be reshuffled once placed inside the container, cannot be placed below an existing parcel, the corners of the parcel must be level, and the base must be flat (Fig. \ref{fig:assumptions}).

In parallel to the real environment, a physics-enabled simulation environment is also created which is used to understand and visualize the different packing strategies. This environment is used for training the RL algorithm, since the physical setup is too slow to be used for training. The goal of the algorithm is to pack a finite stream of incoming boxes (cuboidal, non-deformable, with arbitrary dimensions) into a set of containers. There is no limit on the number of containers that can be used; however, the objective is to minimise the number of non-empty containers after the entire stream of boxes has been packed. In this paper, we work with synthetic data sets that can be publicly shared. Therefore, the optimal number of containers (bins) is known, and the competitiveness ratio $c$ as per Definition 1 can be empirically computed. We also track the volume fraction of the first $Opt(I)$ bins as a secondary objective for all competing algorithms, where $Opt(I)$ is the optimal number of containers as defined in the introductory description.

\section{Algorithms} \label{sec:algorithm}

\begin{figure}[b]
\centering
\includegraphics[width=0.99\linewidth]{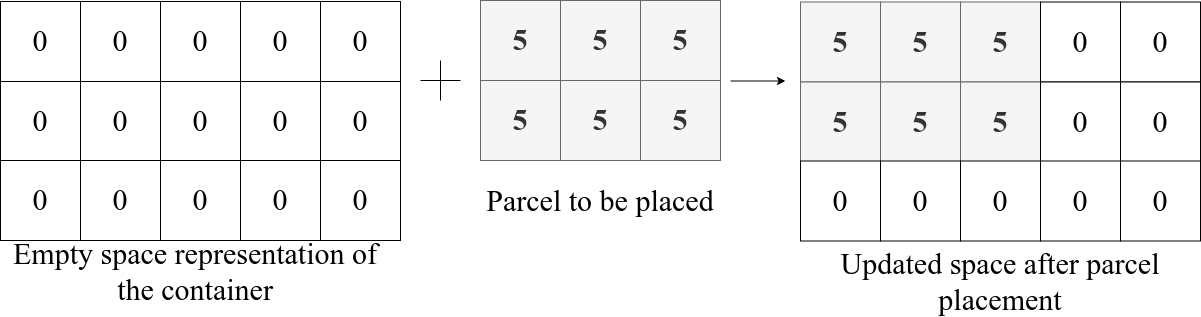}
\caption{The figure shows an empty container of $5 \times 3 \times 5$ on the left. A box of size $3 \times 2 \times 5$ as shown in the center is to be placed inside the container. The rightmost image shows the updated space after placing the box in the top-left corner. }
\label{fig:spacerepresentation}
\end{figure}

We describe the algorithms used for solving the RT-3D-BPP in this paper, including the baseline algorithms from prior literature, a new heuristic developed for this work, and the RL-based approach. All the algorithms described here, model the containers as 2D arrays, representing their state when looking from a top-down perspective. The containers are discretized into a rectangular grid of uniform size (typically 1$\times$1 cm). The tuple $(i,j)$ represents the physical location in the container, while the value $h_{i,j}$ of the grid cell $(i,j)$ represents the height to which boxes have been filled at that location. Each container has dimensions $L$, $B$, and $H$ along the $i$, $j$, and $h$ directions respectively, as shown in Fig. \ref{fig:spacerepresentation}. Placement of a box changes the state of the container as shown, and must not exceed the dimensions $(L,B,H)$ of the container. If there is more than one container open simultaneously, boxes must be fully contained within a single container.

\subsection{Baseline heuristics from literature}
	
We select three standard algorithms that occur in prior literature in various forms \cite{martello2007algorithm}, \cite{den2003note}, \cite{Mahvash2017}, \cite{justesen2018automated}.

\textit{First Fit} places boxes in the first found feasible location (defined by the robot-implementability constraints), scanning row-by-row from the top left of the container from the perspective of Fig. \ref{fig:spacerepresentation}. If no feasible locations are found in the currently available containers, the orientation of the box is changed and the search is executed again. If this check also fails, a new container is opened and the box is placed there. First Fit is the fastest and simplest of the search-based algorithms as it does not have to explore the whole space to find the placement position. 

\textit{Floor building} attempts to pack the container layer-by-layer, from the floor up. In effect, the heuristic places boxes at the lowest (in terms of coordinate $h$) feasible location in the container. Rules for changing the orientation and opening new containers remain the same as for First Fit. Floor building performs very well when boxes in the incoming stream are of similar height, because the newly created surface is as smooth as the base. When the parcels are of varying heights, the solution quality deteriorates because it creates rough surfaces. The algorithm also requires searching through all possible locations and orientations before placing each parcel, leading to slow decision-making.

\textit{Column building} is the vertical complement of floor building, where the algorithm attempts to build towers of boxes with the highest feasible $h$ coordinate in the container. Column building performs very well when the incoming boxes are sorted in decreasing order of their volume. Broadly, column building performs empirically as well as first fit but the overall structure which is created after packing can be unstable, especially for a robot to build.

\subsection{Freshly crafted heuristic: WallE} \label{subsec:WallE}

The key takeaway from the baseline heuristics is that each has its own merits and demerits in specific situations. We also observed expert human packers performing the same job, and learned that they too use a mix of these strategies. Therefore, we attempted to define a heuristic which mimics the desirable characteristics of multiple standard methods. The algorithm is called WallE because it uses incoming boxes to build walls that are of nearly the same height as the neighbouring walls on all four sides. WallE takes the box dimension as input and maintains the state space representation for each container. When a new box arrives, it computes a \textit{stability score} $S$ for each feasible location using the following relationship.

Let us assume that the length and width of the next incoming box is $l$ and $b$ respectively. If this box is placed at location $(i,j)$, it will occupy a total of $l\cdot b$ grid cells from $(i:i+l-1,\,j:j+b-1)$. We first check whether this placement is feasible, according to the rules outlined in problem description. Next, we compute the net variation $G_\mathrm{var}$, which is defined as the \textit{sum of absolute values of differences in cell heights with neighbouring cells around the box}, after the box is placed in the proposed location. Note that $G_\mathrm{var}$ is composed of $l\cdot b$ terms, each corresponding to one bordering cell of the box. If one or more edges of the box are flush with the wall, those quantities are filled by zeroes. Second, we count the number of bordering cells that are higher than the height of the proposed location after loading. Denoted by $G_\mathrm{high}$, this count indicates how snugly the current location packs an existing hole in the container. Finally, we count the number $G_\mathrm{flush}$ of bordering cells that would be exactly level with the top surface of the box, if placed in the proposed location. This indicates how smooth the resulting surface will be.

The stability score definition is given by (\ref{eqn:cost}), where $h_{i,j}$ is the height of the location $(i,j)$ if the box was loaded there. The constants $\alpha_i$ are user-defined \textit{non-negative} parameters (we use $\alpha_1 = 0.75, \alpha_2 = 1, \alpha_3 = 1, \alpha_4 = 0.01, \alpha_5 = 1$ after experimentation; the idea is to ensure all terms are of the same order of magnitude). This score is computed for all locations in the container for both orientations of the box, and the location-orientation combination with the highest score is chosen for placement.
\begin{equation}
S = -\alpha_1\,G_\mathrm{var} + \alpha_2\,G_\mathrm{high} + \alpha_3\,G_\mathrm{flush} - \alpha_4\,(i+j) - \alpha_5\,h_{i,j}
\label{eqn:cost}
\end{equation}	
The structure of (\ref{eqn:cost}) includes characteristics of floor building (penalty on height $h_{i,j}$), first fit (penalty on location $i+j$), as well as an emphasis on smooth surfaces (first three terms), which infuses some wall building tendencies if the resulting placement is tall but smooth on top. While WallE has reasonably good performance as discussed subsequently in results, the definition of $S$ is somewhat arbitrary, and invariant with the characteristics of boxes seen in each data set. It is also dependent on the values of $\alpha_i$. On the other hand, RL has the potential to learn more flexible policies based on training, rather than on hand-tuned parameters. This aspect is explored in the rest of the paper, while WallE provides stronger competition to the RL algorithm than the baseline heuristics.
	
\subsection{Deep Reinforcement Learning - PackMan} \label{subsec:PackMan}
	
The RT-3D-BPP task can be modelled as a Markov Decision Process ($\mathcal{S}, \mathcal{A}, \mathcal{T}, \mathcal{R}, \gamma$), where $\mathcal{S}$ is the set of possible current states, $\mathcal{A}$ denotes the decision or action space, $\mathcal{T}$ represents transition probabilities from one combination of state and action to the next, $\mathcal{R}$ denotes the rewards, and $\gamma$ is the discount factor for future rewards. The Markovian assumption is valid since the current state of the container(s) encapsulates information from all previous loading steps. The environment tracks the state of the containers and the sequence of incoming boxes on the conveyor. Each step is defined by the placement of an incoming box, and an episode terminates when the last box in the sequence is placed. The definitions of states, actions, and rewards are described in detail later in this section.

\textbf{Selection of methodology: }
The obvious approach using reinforcement learning for this problem is to train a policy-based method that computes the optimal location and orientation for the next box, given the current state of the container and (optionally) information about other upcoming boxes. However, we consider this option to be infeasible for the problem at hand. First, the number of possible actions in each step is the product of the number of grid cells and the number of orientations, which can easily exceed ten thousand (see results section). Second, each action is based on the current state of the containers and the dimensions of the next box in the sequence. Understanding the geometric implications of these two inputs in order to predict future rewards is a very complex task. We experimented extensively with policy-based approaches for this task, using methods such as DDPG \cite{ddpg_modified}, but without success. The key obstacle to effective learning was the fact that the optimality of a chosen location and orientation varies sharply (almost discontinuously) with small changes in the inputs. For example, placing a box flush with the wall would lead to good packing, but placing it just 1 cm away from the wall could result in the narrow gap becoming unusable. As a result, we divide the decision-making into two steps: (i) selection of feasible location/orientation combinations using basic rules, and (ii) a value-based RL algorithm for choosing one of the suggested options. We denote the combined procedure by the name PackMan, for brevity.

\begin{figure}[!b]
	\centering
	\includegraphics[width=0.9\linewidth]{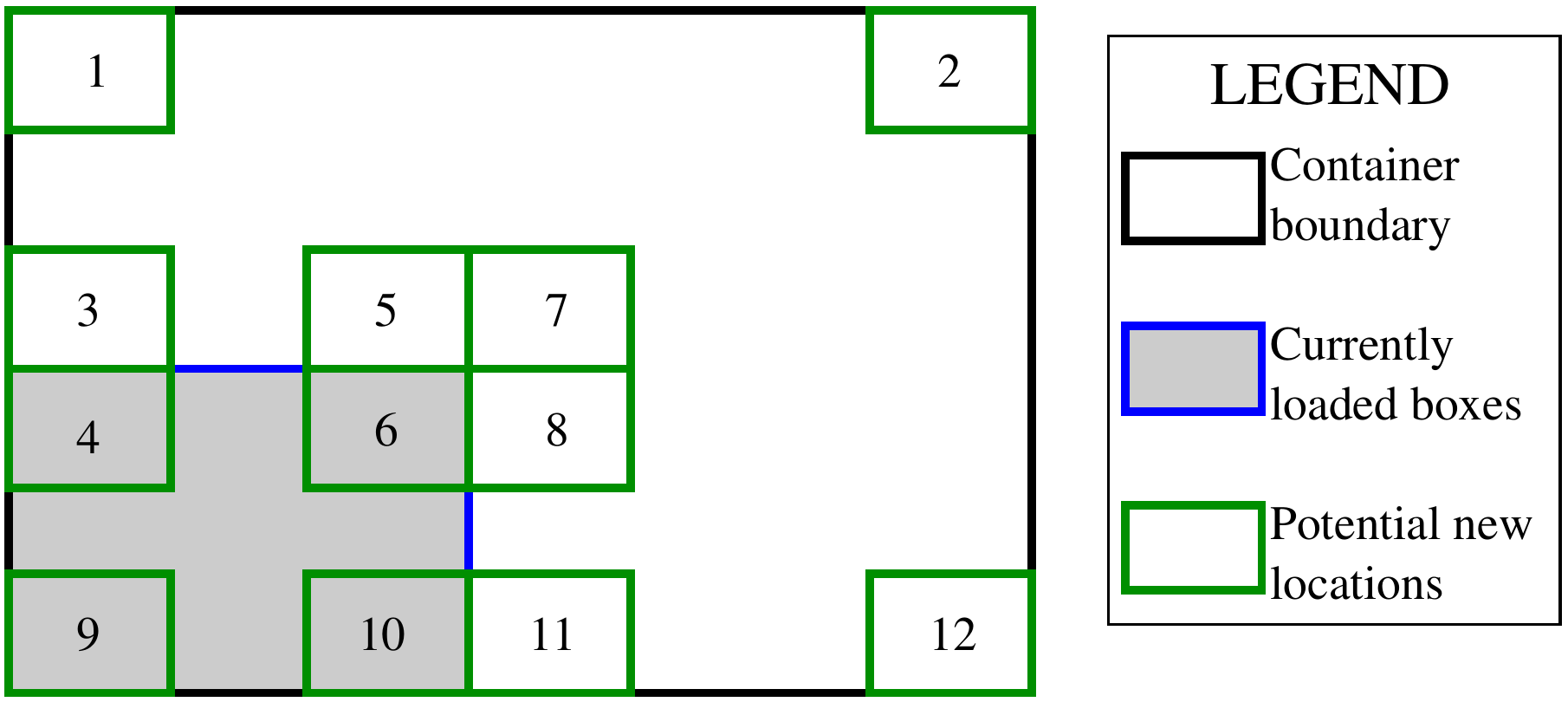} 
 	\caption{Schematic of selective search algorithm. Given a current layout of boxes in the container (shaded in the figure), the algorithm only suggests locations (marked 1 through 12) where a corner of the current box coincides with a corner of the container, or with a corner of the currently loaded box structure.}
 	\label{fig:corners}
\end{figure}

\textbf{Conceptual approach: }
The first step in PackMan is to identify a set of eligible locations and orientations. Given the very large number of grid positions may be placed and the infeasibility of computing the value function for thousands of such options, we narrow the list of eligible locations/orientations using the following logic, with an accompanying schematic shown in Figure \ref{fig:corners}. We note that it is highly unlikely that the optimal location for an upcoming box will be in the middle of a flat surface, or partway along a wall. It is much more likely to coincide with a corner of the container, or with the edges of previously packed boxes within the container. Therefore, given the current container state with a certain number preloaded boxes (shaded portion in Figure \ref{fig:corners}), the location selection procedure only suggests options that coincide with corner locations (for example, the 12 options marked in Figure \ref{fig:corners}). These could be on the floor of the container or on top of other boxes. The same procedure is repeated for both possible orientations of the upcoming box, and the concatenated shortlist is given to the RL agent for choosing the best option.

In the second step, a Deep Q Network \cite{mnih2015human} is used to select one location/orientation combination from the set provided to it. The options are presented in the form of potential future states of the container, were the box to be loaded in the relevant location and orientation. The box is loaded according to the chosen option, the container state is updated, and the environment proceeds to the next box. Skipping the box is not allowed. Instead, if there is no feasible location available for the current box, a new container is opened. The advantages of this approach are, (i) the goodness of each option is learnt using the DQN approach, rather than a hand-tuned score as in WallE, and (ii) the agent does not need to learn the correlation between current box dimensions and the state of the container, but instead only chooses the best out of the set of potential future states, as explained in subsequent text.

\textbf{State definition: }
The state representation for PackMan builds upon the one depicted in Figure \ref{fig:spacerepresentation}, where we look at a top-down view of $T$ containers (placed next to each other) and the value of each cell $(i,j)$ is the total height of stacked boxes in that cell. The number $T$ of containers included in the representation is large enough to fit all possible box sequences, even with poor volume efficiency. The current state of the containers is thus a 2D array with a total of $T\cdot L\cdot B$ grid cells, with the number of cells along each axis dependent on the arrangement of containers. In most of the experiments in this paper, we use $T=16$ containers each of dimension $45\times 80$ placed in a row, resulting in a container state dimension of $45 \times 1280$. 

In order to keep the RL agent independent of the size of the input, we encode the container state into a fixed-size representation $\bar{x}$. This representation could be learnt by an autoencoder, or one can directly use pooling functions. In this work, we use three different pooling functions to reduce any input size to vectors of size 144 each: (i) average pooling, (ii) max pooling, and (iii) the difference between max pooling and min pooling. The vector $\bar{x}$ (of size $3\times 144=432$) is expected to indicate to the RL agent the average height of the underlying receptive fields, as well as its smoothness. The step size of pooling ensures disjoint receptive fields for each element of the vector. In addition, we define two more input channels. A vector $\bar{y}$ encodes the height of the bordering cells of the proposed placement location of the box, in order to indicate how well it fits with the surrounding cells. We use a size of 144 units for this representation, with borders smaller than 144 units (depending on perimeter of the box) padded by trailing zeroes, and borders larger than 144 units populated using constant-skip sampling. Finally, a vector $\bar{z}$ is a one-hot encoding of the receptive field that the currently proposed location belongs to. Together, $(\bar{x},\bar{y},\bar{z})$ define the DQN input.

\begin{figure}[!t]
	\centering
	\includegraphics[width=0.8\linewidth]{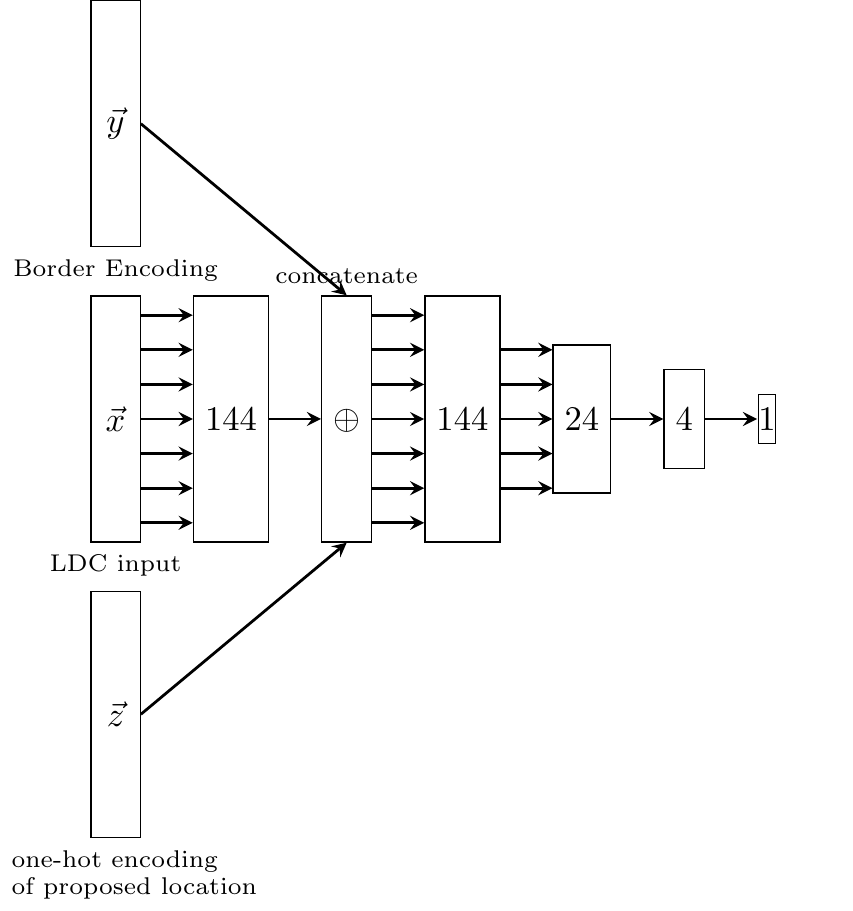} 
 	\caption{Network architecture for the DQN agent. Input images of any shape (in the present instance, $45 \times 1280$) are preprocessed to form three constant-sized vectors $\bar{x}$, $\bar{y}$, and $\bar{z}$ as shown. This is followed by a set of dense layers and channel concatenation to produce a scalar prediction of q-value.}
 	\label{fig:network}
\end{figure}

\textbf{Architecture:}
The network architecture used for the DQN agent is shown in Figure \ref{fig:network}. The current state of the container (denoted by LDC input, where LDC is short for Long Distance Container) is encoded and flattened by the first input $\bar{x}$ of 432 units, which is followed by a dense layer of 144 units. It is then concatenated with the border information $\bar{y}$ and one-hot encoding $\bar{z}$ of the proposed location before passing through the second layer of size 144., followed by a third layer with 24 hidden units, a fourth layer with 4 units, and scalar output neuron representing the q-value. The network is implemented using $\mathtt{keras}$ in $\mathtt{Python}$ 3.6. We use $\mathtt{tanh}$ activations and train using stochastic gradient descent, a learning rate of $0.001$, and momentum of $0.5$.

\textbf{Rewards and training: }
There is no obvious step reward in this problem, since quantitative terms such as volume filled so far are out of our control (order of boxes cannot be changed). Reward shaping using the score from WallE is a possibility, but this would be accompanied by the difficulty of tuning the parameters $\alpha_i$, which we wish to avoid. Therefore, we choose not to give a traditional step reward, and instead provide a discounted version of the terminal reward as a proxy for the step reward. The terminal reward itself is defined by,
\begin{equation}
\zeta = \frac{V_{packed}}{T_{used}\cdot L \cdot B \cdot H} - \tau, \label{eq:terminal}
\end{equation}
where the first term is the packing fraction for the whole sequence (assuming that the number of occupied containers at the end of the episode is $T_{used}$), and $\tau$ is the average packing fraction over all episodes since the start of training. The terminal reward encourages continuous improvement during the training process, an idea analysed in prior work\footnote{self-citation redacted for anonymity}. If there are $N$ boxes in a given episode, then the step reward at time step $t$ is given by,
\begin{equation*}
r_t = \rho^{N-t}\zeta,
\end{equation*}
where $\rho = 0.99$ is a discount factor for the terminal reward. The entire episode is inserted into the replay buffer after completion and computation of the rewards for each step. The logic for using this reward definition (inspired by on-policy algorithms) is (i) to provide a reward based on the results of the entire sequence of decisions, and (ii) to speed up training by avoiding the sparse nature of pure terminal reward approaches.

Training is carried out with a mean-squared error loss with respect to the following relation with $\gamma = 0.75$,
\begin{equation*}
Q(s_t, a_t) = (1-\gamma)r_t + \gamma Q(s_{t+1}, a_{t+1}) , 
\end{equation*}
where the left-hand side is network output and the right-hand side is the target value, produced by a target network (cloned from the online network after every 10 training runs). Training is carried out after every episode with a batch size of 256 (the approximate number of boxes per episode in the training data).

\section{Results} \label{sec:results}

We train PackMan using synthetically generated data sets, containing boxes of randomly generated dimensions. However, we ensure that the dimensions match up such that each container can be filled completely (100\% fill fraction). Each data set consists of 10 containers worth of boxes ($Opt(I)=10$), with the number of boxes ranging between 230 and 370 per episode. The order of upcoming boxes is not known to the algorithms, apart from $n=2$ boxes after the current one. The episode terminates when all boxes are packed, or if the algorithm is unable to pack all boxes in a maximum of 16 containers. The terminal reward (\ref{eq:terminal}) is computed based on the number of containers required in each episode.

\begin{figure}[b!]
	\centering
    \includegraphics[width=0.99\linewidth]{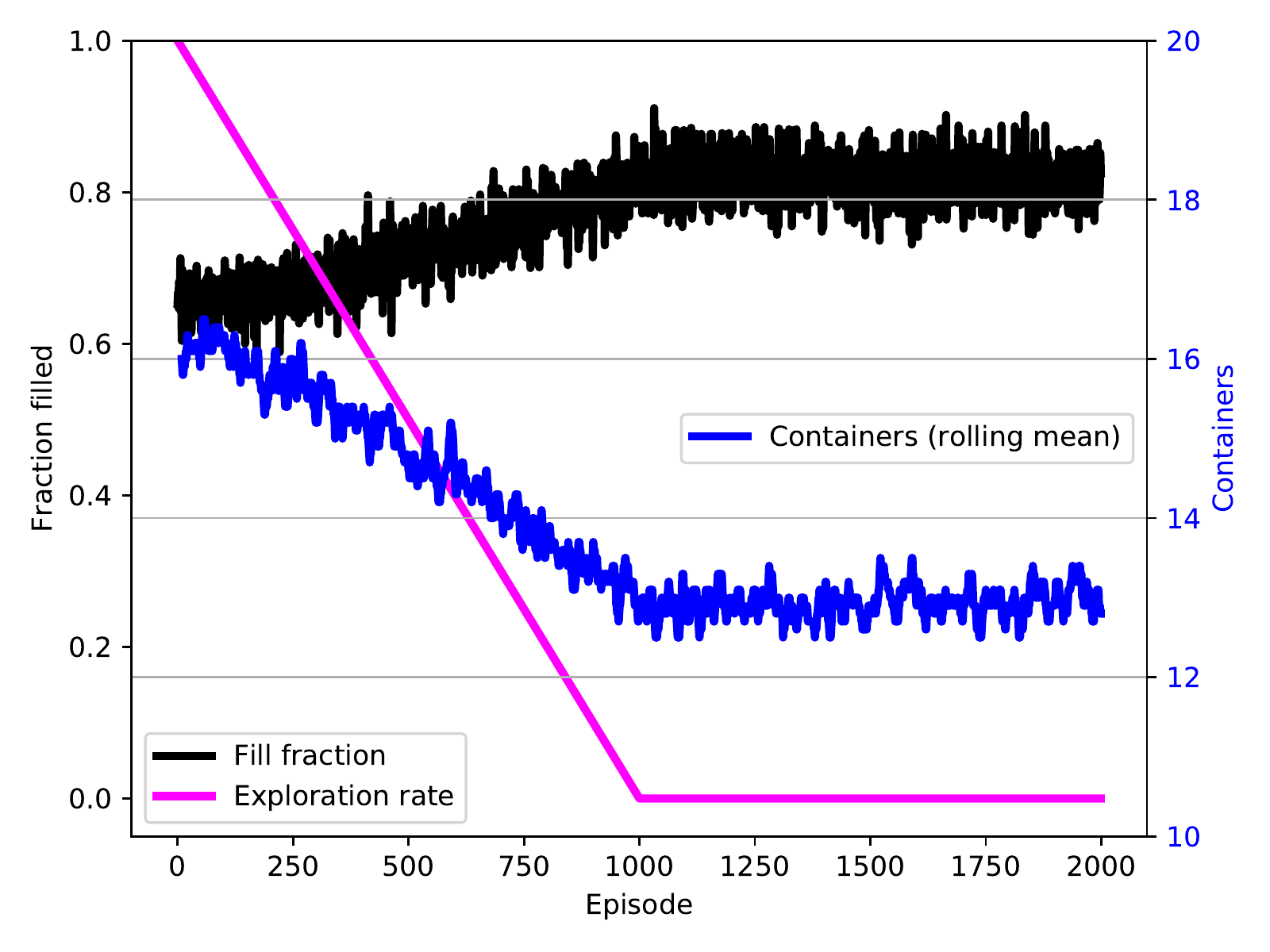}
	\caption{Fill percentages of first 10 bins, and total bins used, during the course of RL training.}
	\label{fig:training}
\end{figure}

\textbf{Training: }
Fig. \ref{fig:training} shows the improvement in packing efficiency over 2000 episodes of training using $\epsilon$-greedy exploration, with $\epsilon$ decreasing linearly from 1 to 0 in 1000 episodes. The algorithm trains for the last 1000 episodes without exploration, because we found that random decisions early in an episode greatly affect the terminal reward, significantly slowing down training. The initial packing efficiency of approximately 65\% improves steadily to 82\% over 1100 episodes, and stays stable afterwards. The number of bins used decreases from just above 16 (we tag episodes that do not terminate in 16 bins with a bin count of 17) to just under 13.

\textbf{Comparison with baselines: }
Table \ref{table:results} compares the algorithms on the competitiveness ratio metric ($c$), the time taken per loading decision, average packing efficiency, and the fraction of test instances in which a given algorithm returned the best packing efficiency. Advanced Harmonic (AH) is known to have a theoretical upper bound of $c=1.58$, although this is with unconstrained rotation. Empirical results for robot-stackable algorithms show that PackMan has the best empirical ratio of $1.29$, averaging $T_{used}=12.9$ bins compared to $Opt(I)=10$. It also has the highest average packing fraction. While the difference in packing fractions is small, further investigation revealed that this was because there was significant variation among the instances, with some box streams favouring one algorithm over the others. The fact that PackMan returns the best efficiency in 57\% of test cases implies that it retains a significant advantage over other algorithms across a variety of instances.
	
	\begin{table}[t!]
		\centering
		\begin{tabular}{|p{2.0cm}|p{1.2cm}|p{1.2cm}|p{1.2cm}|p{0.6cm}|}
			\hline
			
			Algorithm & Comp. ratio $c$  & Time per box (sec) & Avg. pack & Best Pack \\ \hline
			AH                    & 1.58 & -               & -      & -   \\ 
			Floor building        & 1.52 & 0.0002          & 81.0\% & 05\% \\ 
			Column build       & 1.46 & \textbf{0.0001} & 81.0\% & 06\% \\ 
			First Fit             & 1.47 & 0.0002          & 81.3\% & 07\% \\ 
			WallE                 & 1.41 & 0.0106          & 81.8\% & 25\% \\ 
			PackMan      & \textbf{1.29} & 0.0342 & \textbf{82.8\%} & \textbf{57\%} \\        \hline
		\end{tabular} 
		\vspace{1mm}
		\centering
		\caption{Comparison of results on 100 episodes of test data. AH \cite{balogh2017new} has a theoretical competitive ratio of 1.58, and was not tested empirically. Other algorithms have empirical results. While PackMan requires the largest time per inference, it also has the lowest competitive ratio, highest average packing fraction, and largest proportion of episodes where it had the best packing fraction.}
		\label{table:results}
	\end{table}  
	
\begin{figure}[!t]
	\centering
	\includegraphics[width=0.99\linewidth]{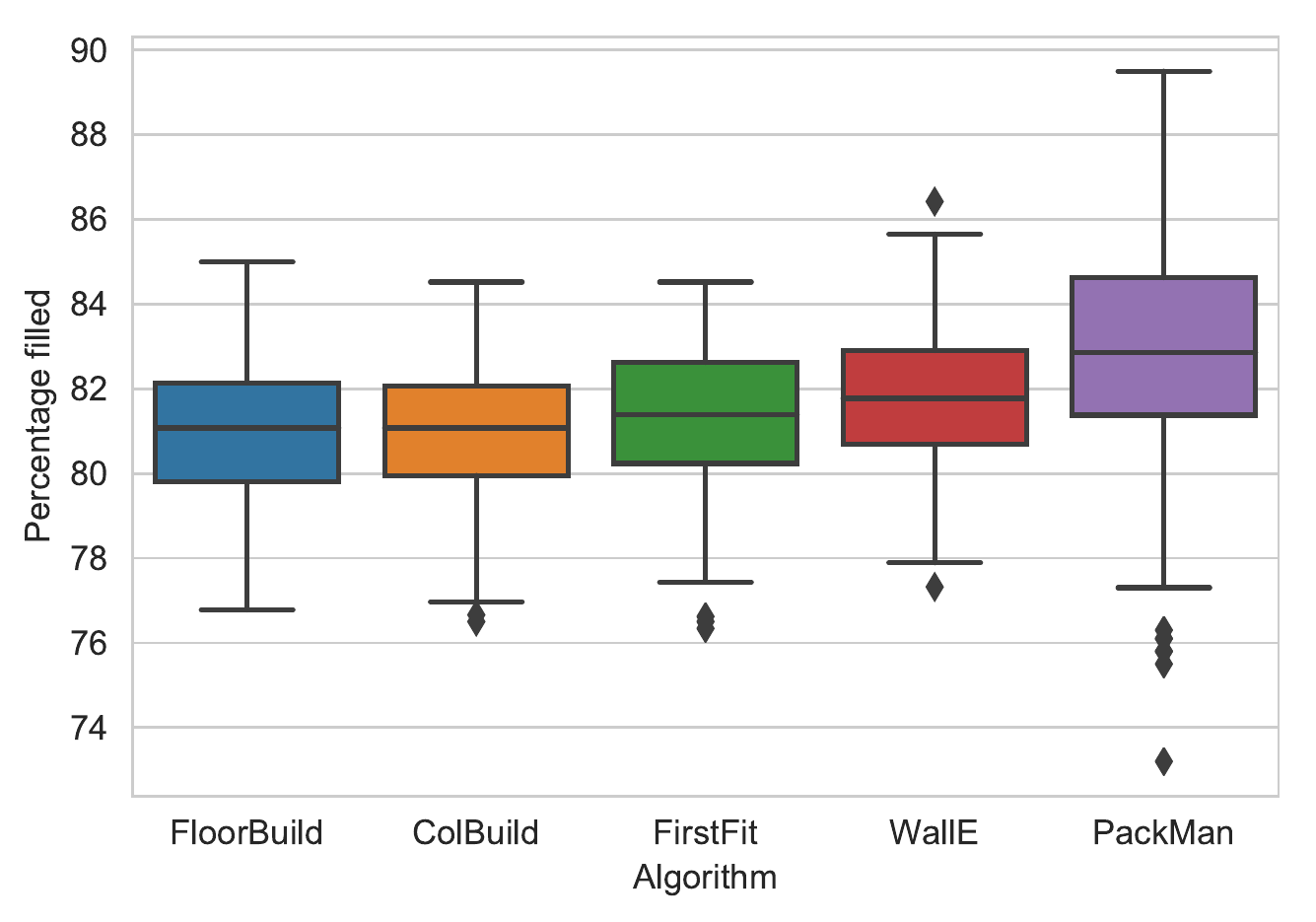} 
 	\caption{Comparison of empirical fill rates for all 5 algorithms, over 100 test data sets.}
 	\label{fig:comparison}
\end{figure}

The box whiskers plot shown in Figure \ref{fig:comparison} illustrates the differences between the algorithms. While floor building and column building have nearly identical results for the test data sets, WallE returns the best results among the heuristics. This is a result of its balanced approach to box placement, without a singular emphasis on floor or column building. The median packing efficiency for PackMan is clearly higher than all the heuristics, but it has a larger spread in the outliers. However, none of the PackMan test runs resulted in a competitive ratio $c$ higher than 1.4 (or 14 bins).
	
\begin{figure}[!t]
\centering
\includegraphics[width=0.99\linewidth]{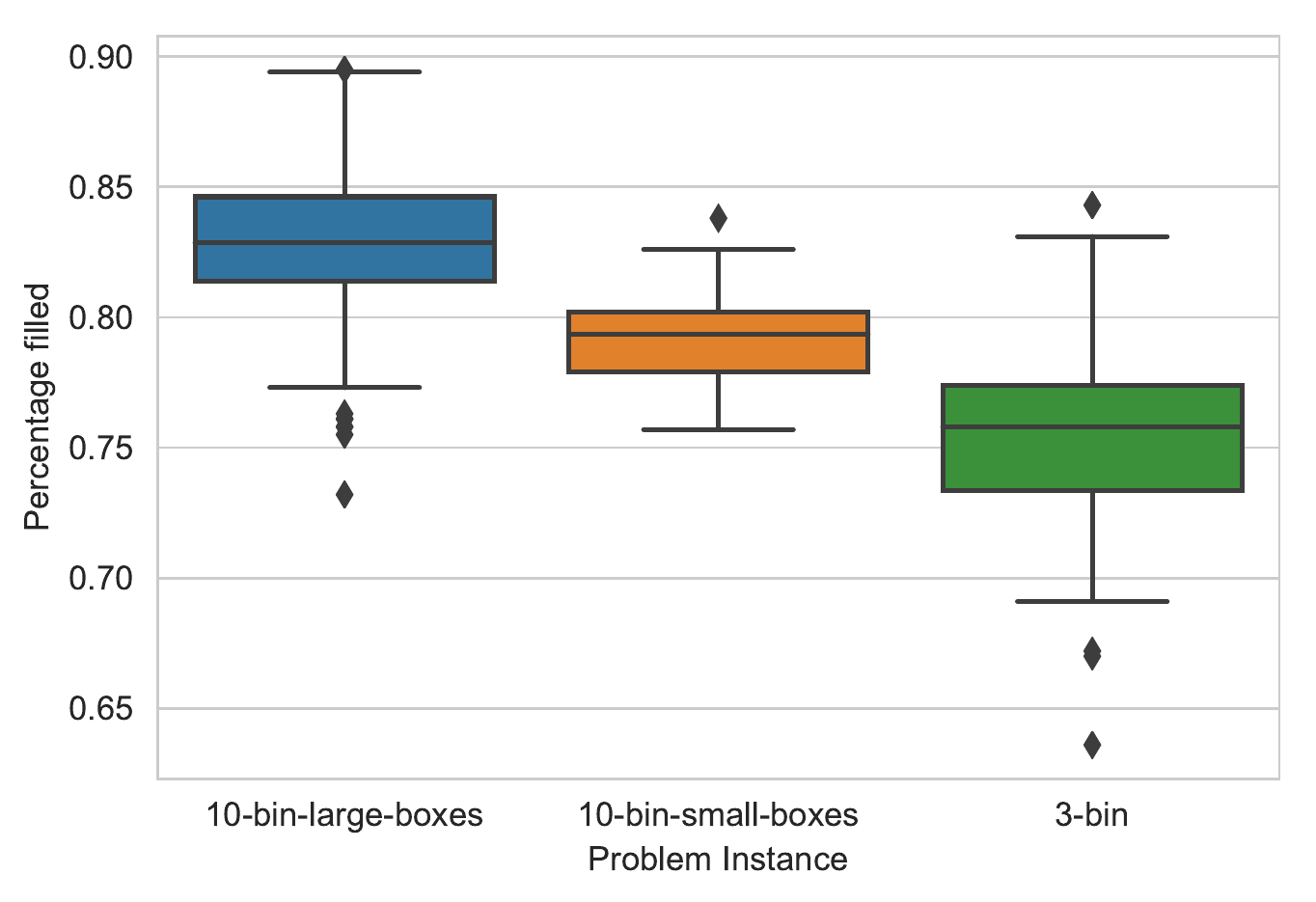} 
\caption{Comparison of empirical fill rates for three different problem instances, over 100 test data sets each.}
\label{fig:transfer}
\end{figure}

\begin{figure}[!t]
	\centering
	\includegraphics[width=0.6\linewidth]{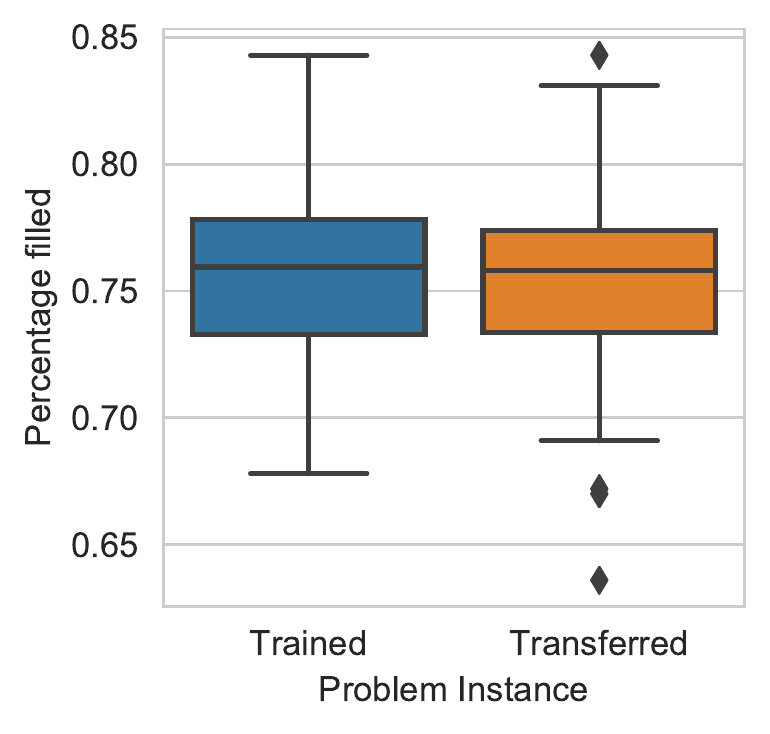} 
 	\caption{Comparison of empirical fill rates for the 3-bin data set, with training on 3-bin data and with weights transferred directly from 10-bin data.}
 	\label{fig:self-transfer}
\end{figure}

\textbf{Transfer learning capability: }
As mentioned earlier in the paper, the goal of the present approach is to train an RL that works without retraining on other problem instances, whether generated using a different distribution of box dimensions, or a different scale of inputs. Figure \ref{fig:transfer} plots the packing efficiency in two additional types of instances. The left-most box plot is a repeat of the PackMan results from Figure \ref{fig:comparison}. The box plot in the middle is PackMan's efficiency when boxes have smaller dimensions on average. The underlying random distribution is over smaller dimensions. Even though the total volume is still equivalent to 10 bins, there are roughly double the number of boxes per episode. Since smaller dimensions result in rougher surfaces, the average packing efficiency is lower, but the degradation is small. The right-most box plot in Figure \ref{fig:transfer} plots the results on a data set with $Opt(I)=3$ bins. We allow a maximum of 6 bins in this case, with a raw input size of $45\times(80\times 6)$ pixels. It is still mapped to the same vector inputs $(\bar{x},\bar{y},\bar{z})$, and the same model learnt for 16 bins is used without retraining. While the packing efficiency is even lower, this is likely to be a characteristic of these data sets. Figure \ref{fig:self-transfer} shows that the efficiency when the model is trained on 3-bin data sets is nearly the same as that of the model transferred from the 10-bin data set.

\textbf{Insights from real-world experiments: }
The final version of PackMan was deployed on a real robotic system to understand the challenges. The most critical challenge in deployment was updating the algorithm's belief state according to the containers' actual configuration, which are subject to sensor and actuator noise. Because of this, the parcels may not get placed at the exact location dictated by the algorithm. 
We observed that parcels collide inside the container during placement due to these errors. To avoid collision we increased the grid size in the belief state of the algorithm and left a space of around 1-2 cm between boxes. Eliminating this space is possible only when the belief state of the algorithm is updated using millimeter-level accurate real-time measurements. Another challenge was estimation of box orientation and dimension while picking. Any error at this point is carried forward and causes difference in actual and belief state. 


\section{Conclusion and Future Work} \label{sec:results}
This paper presented PackMan, a two-step approach to tackle the online 3D Bin Packing Problem using a search heuristic followed by reinforcement learning. We tested our approach on multiple randomly generated datasets. The experiments demonstrated that PackMan does better than the state-of-the-art online bin packing heuristics and generates robot packable solutions which are essential for real world deployment. We also showed that PackMan can be used on different box distributions and scales of input without re-training.
While PackMan can be used in real-time with an inference time under 40 ms, we foresee potential for further improvements. We would like to improve the RL strategy by reworking the input layers of the network to transmit more information about the container layout. There is also the possibility of looking at direct policy-based outputs through more advanced architectures. The hardware capabilities of the robotic loading system are being improved in parallel with the algorithmic improvement efforts.

\bibliographystyle{aaai}
\bibliography{ref_hk}

\end{document}